\documentclass{article}

\usepackage{ijcai13}

\newtheorem {definition}{Definition}
\newtheorem {proposition}{Proposition}

\newtheorem {theorem}{Theorem}

\newtheorem {example}{Example}

\usepackage{amsmath}
\usepackage{amssymb}
\usepackage{rotating}

\begin{document}

\title{Probability Aggregates in Probability Answer Set Programming}

\author{ Emad Saad \\
emsaad@gmail.com
}

\maketitle

\begin{abstract}
Probability answer set programming \cite{Saad_B,Saad_EHPP,Saad_DHPP} is a declarative programming that has been shown effective for representing and reasoning about a variety of probability reasoning tasks  \cite{Saad_MDP,Saad_Learn_Sense,SaadPlan,Saad_Sensing,SaadSSAT}. However, the lack of probability aggregates, e.g. {\em expected values}, in the language of disjunctive hybrid probability logic programs (DHPP) \cite{Saad_DHPP} disallows the natural and concise representation of many interesting problems. In this paper, we extend DHPP to allow arbitrary probability aggregates. We introduce two types of probability aggregates; a type that computes the expected value of a classical aggregate, e.g., the expected value of the minimum, and a type that computes the probability of a classical aggregate, e.g, the probability of sum of values. In addition, we define a probability answer set semantics for DHPP with arbitrary probability aggregates including monotone, antimonotone, and nonmonotone probability aggregates. We show that the proposed probability answer set semantics of DHPP subsumes both the original probability answer set semantics of DHPP \cite{Saad_DHPP} and the classical answer set semantics of classical disjunctive logic programs with classical aggregates \cite{Recur-aggr}, and consequently subsumes the classical answer set semantics of the original disjunctive logic programs \cite{Gelfond_B}. We show that the proposed probability answer sets of DHPP with probability aggregates are minimal probability models and hence incomparable, which is an important property for nonmonotonic probability reasoning.

\end{abstract}

\section{Introduction}

Probability answer set programming \cite{Saad_B,Saad_EHPP,Saad_DHPP} is a declarative programming framework which aims to solve hard search problems in probability environments, and shown effective for probability knowledge representation and probability reasoning applications. It has been shown that many interesting probability reasoning problems are represented and solved by probability answer set programming, where probability answer sets describe the set of possible solutions to the problem. These probability reasoning problems include, but not limited to, reasoning about actions with probability effects and probability planning \cite{SaadPlan}, reinforcement learning in MDP environments \cite{Saad_MDP}, reinforcement learning in POMDP environments \cite{Saad_Learn_Sense}, contingent probability planning \cite{Saad_Sensing}, and Bayesian reasoning \cite{SaadSSAT}.  However, the unavailability of probability aggregates, e.g. {\em expected values}, in the language of probability answer set programming \cite{Saad_B,Saad_EHPP,Saad_DHPP} disallows the natural and concise representation of many interesting problems. This requires probability answer set programs to be capable of representing and reasoning in the presence of {\em probability aggregates}. The following stochastic dietary problem illuminates the need for probability aggregates.


\begin{example}

Suppose we have three kinds of food: beef, fish, and turkey, where the amounts of vitamins of $A$, $B$, and $C$ per unit of each of these food are uncertain. Two scenarios are available for each amount of units of vitamins for each unit of food. The amounts of units of vitamins $A$, $B$, and $C$ per unit of beef are believed to be $(60, 10, 20)$ with $(0.7, 0.6, 0.8)$ probability and $(50, 8, 15)$ with $(0.3, 0.4, 0.2)$ probability. Per unit of fish, the amounts of units of vitamins are believed to be $(8, 15, 10)$ with $(0.8, 0.5, 0.4)$ probability and $(11, 18, 13)$ with $(0.2, 0.5, 0.6)$ probability. Per unit of turkey, the amounts of units of vitamins are believed to be $(60, 15, 20)$ with $(0.8, 0.7, 0.9)$ probability and $(55, 20, 25)$ with $(0.2, 0.3, 0.1)$ probability.

Assume each kind of food is available in packages of $1$ or $2$ units, presented by the predicate $pckg(F,N, S)$, where $F$ is a food, $N$ is the number of units of the food $F$, and $S$ is the scenario in which the package is selected. We use $units(F, V, U, S): P$ to represent a unit of food $F$ has $U$ units of vitamin $V$ with probability $P$ in a scenario $S$. The minimum daily requirement of vitamins $A$, $B$, and $C$ is $230$, $75$, and $95$ units, respectively.

The target is to find combinations of units of food that meet the minimum daily requirement of each vitamin. This requires finding the {\em expected value} of units of vitamins for each vitamin collected from each available food in every possible scenario, and compare this expected value with the minimum daily requirement of each vitamin.

This probability optimization problem can be represented by a disjunctive hybrid probability logic program with probability answer set semantics, DHPP \cite{Saad_DHPP}. DHPP is an expressive probability answer set programming framework \cite{Saad_B,Saad_EHPP,Saad_DHPP} that allows disjunctions in the head of rules. We assume that atoms appearing without annotations, in DHPP programs, are associated with the annotation $[1,1]$, and annotated atoms of the form $A : [\alpha,\alpha]$ are simply represented as $A:\alpha$. The DHPP program representation, $\Pi = \langle R, \tau \rangle$, of the stochastic dietary problem, is given as follows, where $\tau$ is any arbitrary assignment of disjunctive p-strategies and $R$ contains rules of the form:
{\small
\[
\begin{array}{rrr}
food(beef)  \leftarrow &  \hspace{0.7cm} food(fish)  \leftarrow &  \hspace{0.7cm} food(turkey)  \leftarrow
\end{array}
\]
\[
\begin{array}{lcllcl}
units(beef, a, 60, s_1):0.7  & \leftarrow &   units(beef, b, 10, s_1):0.6  & \leftarrow &
\\
units(beef, a, 50, s_2):0.3  & \leftarrow &   units(beef, b, 8, s_2):0.4  & \leftarrow &
\\
units(fish, a, 8, s_1):0.8  & \leftarrow &   units(fish, b, 15, s_1):0.5  & \leftarrow &
\\
units(fish, a, 11, s_2):0.2  & \leftarrow &   units(fish, b, 18, s_2):0.5  & \leftarrow &
\\
units(turk, a, 60, s_1):0.8  & \leftarrow &   units(turk, b, 15, s_1):0.7  & \leftarrow &
\\
units(turk, a, 55, s_2):0.2  & \leftarrow &   units(turk, b, 20, s_2):0.3  & \leftarrow &
\\
units(beef, c, 20, s_1):0.8  & \leftarrow &   units(beef, c, 15, s_2):0.2  & \leftarrow &  \\
units(fish, c, 10, s_1):0.4  & \leftarrow &   units(fish, c, 13, s_2):0.6  & \leftarrow &  \\
units(turk, c, 20, s_1):0.9  & \leftarrow &   units(turk, c, 25, s_2):0.1  & \leftarrow & \\
 \end{array}
\]
\[
\begin{array}{lcl}
pckg(F, 1, S) \;\;\vee\;\; pckg(F, 2, S) & \leftarrow & food(F) \\
nutr(F, V, U \times N, S) : P & \leftarrow & units(F, V, U, S) : P, \\ && pckg(F, N, S)
\end{array}
\]
\[
\begin{array}{r}
expected(a, U_1*P_1 + U_2*P_2 + U_3*P_3 + U_4*P_4 + U_5*P_5 \\ + U_6*P_6)   \leftarrow
                        nutr(beef, a,  U_1, s_1): P_1, \\ nutr(beef, a, U_2, s_2): P_2,
                        nutr(fish, a, U_3, s_1): P_3, \\ nutr(fish, a, U_4, s_2): P_4,
                        nutr(turk, a, U_5, s_1): P_5,  \\ nutr(turk, a, U_6, s_2): P_6 \\

expected(b, U_1*P_1 + U_2*P_2 + U_3*P_3 + U_4*P_4 + U_5*P_5 \\ + U_6*P_6)   \leftarrow
                        nutr(beef, b, U_1, s_1): P_1, \\ nutr(beef, b, U_2, s_2): P_2,
                        nutr(fish, b, U_3, s_1): P_3, \\ nutr(fish, b, U_4, s_2): P_4,
                        nutr(turk, b, U_5, s_1): P_5, \\ nutr(turk, b, U_6, s_2): P_6 \\

expected(c, U_1*P_1 + U_2*P_2 + U_3*P_3 + U_4*P_4 + U_5*P_5 \\ + U_6*P_6)   \leftarrow
                        nutr(beef, c, U_1, s_1): P_1, \\ nutr(beef, c, U_2, s_2): P_2,
                        nutr(fish, c, U_3, s_1): P_3, \\ nutr(fish, c, U_4, s_2): P_4,
                        nutr(turk, c, U_5, s_1): P_5,\\ nutr(turk, c, U_6, s_2): P_6
\end{array}
\]

\[
\begin{array}{lcl}
\Gamma  & \leftarrow & not \; \Gamma, \; expected(a, X), \; X < 230 \\
\Gamma  & \leftarrow & not \; \Gamma, \; expected(b, X), \; X < 75 \\
\Gamma  & \leftarrow & not \; \Gamma, \; expected(c, X), \; X < 95
\end{array}
\]
}
The last three rules in the above DHPP program representation of the stochastic dietary problem guarantee that only probability answer sets with sufficient supply of vitamins are generated.
\label{ex:DHPP_encode}
\end{example}
The DHPP representation of the stochastic dietary problem described in Example (\ref{ex:DHPP_encode}) is fairly intuitive but rather complex, since the rules that represent the expected value of units of vitamins for each vitamin via the predicate $expected(V,E)$, where $E$ is the expected value of units of vitamins for vitamin $V$, contains complex summation that involves $12$ variables. Furthermore, this representation strategy is not feasible in general, especially, in the presence of multiple scenarios for each amount of units of vitamin per unit of food, multiple numbers of vitamins, and multiple types of food, which consequently will lead to very complex rules with very complex summations.

Therefore, we propose to extend the language of DHPP with probability aggregates to allow intuitive and concise representation and reasoning about real-world applications. To the best of our knowledge, this development is the first that defines semantics for probability aggregates in a probability answer set programming framework. DHPP is expressive form of probability answer set programming \cite{Saad_B,Saad_EHPP,Saad_DHPP} that allows disjunctions in the head of rules. It has been shown that; DHPP is capable of representing and reasoning with both probability uncertainty and qualitative uncertainty \cite{Saad_DHPP}; it is a natural extension to the classical disjunctive logic programs, DLP, and its probability answer set semantics generalizes the classical answer set semantics of DLP \cite{Saad_DHPP}; DHPP with probability answer set semantics generalizes the probability answer set programming framework of \cite{Saad_B}, which are DHPP programs with an atom appearing in the heads of rules. Moreover, it has been shown that DHPP is used in real-world applications in which quantitative probability uncertainly need to be defined over the possible outcomes of qualitative uncertainty \cite{Saad_DHPP}.

There were many proposals for defining semantics for classical aggregates in classical answer set programming \cite{Recur-aggr,Smodels-Weight,WFM-Pref,Dis-mono-aggr,Ferraris,FOL-aggr,Pelov}. Among these proposals, \cite{Recur-aggr} is the most general intuitive semantics for classical aggregates in DLP. In \cite {Recur-aggr}, declarative classical answer semantics for classical disjunctive logic program with arbitrary classical aggregates, denoted by DLP$^{\cal A}$, including monotone, antimonotone, and nonmonotone aggregates, was provided. The proposed classical answer set semantics of  DLP$^{\cal A}$ generalizes the classical answer set semantics of aggregate-free DLP. Moreover, classical answer sets of DLP$^{\cal A}$ are subset-minimal \cite{Recur-aggr}, a vital property for nonmonotonic reasoning framework semantics.

The contributions of this paper are the following. We extend the original language of DHPP to allow any arbitrary probability annotation function including monotone, antimonotone, and nonmonotone annotation functions. We define the notions of probability aggregates and probability aggregate atoms in DHPP. We present two types of probability aggregates; the first type computes the expected value of a classical aggregate, e.g., the expected value of the minimum, the second type computes the probability of a classical aggregate, e.g, the probability of sum of values. In addition, we define the probability answer set semantics of DHPP with arbitrary probability aggregates, denoted by DHPP$^{\cal PA}$, including monotone, antimonotone, and nonmonotone probability aggregates. We show that the proposed probability answer set semantics of DHPP$^{\cal PA}$ subsumes both the original probability answer set semantics of DHPP \cite{Saad_DHPP} and the classical answer set semantics of DLP$^{\cal A}$ \cite{Recur-aggr}, and consequently subsumes the classical answer set semantics of DLP \cite{Gelfond_B}. We show that the probability answer sets of DHPP$^{\cal PA}$ are minimal probability models and hence incomparable, which is an important property for nonmonotonic probability reasoning.

\section{DHPP$^{\cal PA}$ : Probability Aggregates Disjunctive Hybrid Probability Logic Programs}

In this section we introduce the basic language of DHPP$^{\cal PA}$, the notions of probability aggregates and probability aggregate atoms, and the syntax of DHPP$^{\cal PA}$ programs.

\subsection{The Basic Language of DHPP$^{\cal PA}$}

Let $\cal L$ denotes an arbitrary first-order language with finitely many predicate symbols, function symbols, constants, and infinitely many variables. A term is a constant, a variable or a function. An atom, $a$, is a predicate in $\cal {B_L}$, where $\cal {B_L}$ is the Herbrand base of $\cal L$. The Herbrand universe of $\cal L$ is denoted by $U_{\cal L}$. Non-monotonic negation or the negation as failure is denoted by $not$. In probability aggregates
disjunctive hybrid probability logic programs, DHPP$^{\cal PA}$, probabilities are assigned to primitive events (atoms) and compound events (conjunctions or disjunctions of atoms) as intervals in ${\cal C}[0,1]$, where ${\cal C}[0, 1]$ denotes the set of all closed intervals in $[0, 1]$. For $[\alpha_1, \beta_1], [\alpha_2, \beta_2] \in {\cal C}[0, 1]$, the \emph{truth order} $\leq_t$ on ${\cal C}[0, 1]$ is defined as $[\alpha_1, \beta_1] \leq_t [\alpha_2, \beta_2]$ iff $\alpha_1 \leq \alpha_2$ and $\beta_1 \leq \beta_2$.

The type of dependency among the primitive events within a compound event is described by \emph{a probability strategy}, which can be a \emph{conjunctive} p-strategy or a \emph{disjunctive} p-strategy. Conjunctive (disjunctive) p-strategies are used to combine events belonging to a conjunctive (disjunctive) formula \cite{Saad_B}. The \emph{probability composition function}, $c_\rho$, of a probability strategy (p-strategy), $\rho$, is a mapping $c_\rho : {\cal C}[0,1] \times {\cal C}[0, 1] \rightarrow {\cal C}[0, 1]$, where the probability composition function, $c_\rho$, computes the probability interval of a conjunction (disjunction) of two events from the probability of its components. Let $M = \{\!\!\{[\alpha_1, \beta_1], \ldots, [\alpha_n, \beta_n]\}\!\!\}$ be a multiset of probability intervals. For convenience, we use $c_\rho M$ to denote $c_\rho ([\alpha_1, \beta_1], c_\rho ([\alpha_2, \beta_2],\ldots, c_\rho([\alpha_{n-1}, \beta_{n-1}],[\alpha_n, \beta_n]))\ldots )$.

A \emph{probability annotation} is a probability interval of the form $[\alpha_1, \alpha_2]$, where $\alpha_1, \alpha_2$ are called probability annotation items. A \emph{probability annotation item} is either a constant in $[0, 1]$ (called \emph{probability annotation constant}), a variable ranging over $[0, 1]$ (called \emph{probability annotation variable}), or $f(\alpha_1,\ldots,\alpha_n)$ (called \emph{probability annotation function}), where $f$ is a representation of a monotone, antimonotone, or nonmonotone total or partial function $f: ([0, 1])^n \rightarrow [0, 1]$ and $\alpha_1,\ldots,\alpha_n$ are probability annotation items.

Let $S = S_{conj} {\cup} S_{disj}$ be an arbitrary set of p-strategies, where $S_{conj}$ ($S_{disj}$) is the set of all conjunctive (disjunctive) p-strategies in $S$. A \emph{hybrid basic formula} is an expression of the form  $a_1 \wedge_\rho \ldots \wedge_\rho a_n$ or $a_1 \vee_{\rho'} \ldots \vee_{\rho'} a_n$, where $a_1, \ldots, a_n$ are atoms and $\rho$ and $\rho'$ are p-strategies. Let $bf_S({\cal B_L})$ be the set of all ground hybrid basic formulae formed using distinct atoms from ${\cal B_L}$ and p-strategies from $S$. If $A$ is a hybrid basic formula and $\mu$ is a probability annotation then $A:\mu$ is called a probability annotated hybrid basic formula.

\subsection{Probability Aggregate Atoms}

A symbolic probability set is an expression of the form $\{ F : [P_1, P_2] \; | \; C \}$, where $F$ is a variable or a function term and $P_1$, $P_2$ are probability annotation variables or probability annotation functions, and $C$ is a conjunction of probability annotated hybrid basic formulae. A ground probability set is a set of pairs of the form $\langle F^g : [P^g_1, P^g_2] \; | \; C^g \rangle$ such that $F^g$ is a constant term and $P^g_1, P^g_2$ are probability annotation constants, and $C^g$ is a ground conjunction of probability annotated hybrid basic formulae. A symbolic probability set or ground probability set is called a probability set term. Let $f$ be a probability aggregate function symbol and $S$ be a probability set term, then $f(S)$ is said a probability aggregate, where $f \in \{$ $val_E$, $sum_E$, $times_E$, $min_E$, $max_E$, $count_E$, $sum_P$, $times_P$, $min_P$, $max_P$, $count_P$  $\}$. If $f(S)$ is a probability aggregate and $T$ is an interval $[\theta_1, \theta_2]$, called {\em guard}, where $\theta_1, \theta_2$ are constants, variables or functions terms, then we say $f(S) \prec T$ is a probability aggregate atom, where $\prec \in \{=, \neq, <, >, \leq, \geq \}$.

\begin{example} The following examples are representation for probability aggregate atoms.
\[
\begin{array}{c}
sum_E \; \{ \; X : [P_1, P_2] \: | \: demand(X): [P_1, P_2]  \; \} \; <  [190, 230] \\
min_P  \; \{ \; \langle 7:[0.2, 0.3] \; | \; a(7, 1):[0.2, 0.3] \rangle,  \\ \langle 2:[0.5, 0.9] \; | \; a(2, 1):[0.5, 0.9] \rangle \; \} \; \ge  [0.45, 0.6]
\end{array}
\]
\label{ex:samples}
\end{example}
Definition (\ref{def:local}) below specifies that every probability aggregate function $f(S)$ has its own set of local variables.

\begin{definition}
Let $f(S)$ be a probability aggregate. A variable, $X$, is a local variable to $f(S)$ if and only if $X$ appears in $S$ and $X$ does not appear in the DHPP$^{\cal PA}$ rule that contains $f(S)$.
\label{def:local}
\end{definition}
For example, for the first probability aggregate atom in Example (\ref{ex:samples}), the variables $X$, $P_1$, and $P_2$ are local variables to the probability aggregate $sum_E$.

\begin{definition}
A global variable is a variable that is not a local variable.
\end{definition}

\subsection{DHPP$^{\cal PA}$ Program Syntax}

A DHPP$^{\cal PA}$ rule is an expression of the form
\[
\begin{array}{r}
a_1:\mu_1 \; \vee \ldots \vee \; a_k:\mu_k \leftarrow A_{k+1}:\mu_{k+1}, \ldots, A_m:\mu_m, \\ not\; A_{m+1}:\mu_{m+1},\ldots, not\;A_{n}:\mu_{n},
\end{array}
\]
where $\forall (1 \leq i \leq k)$ $a_i$ are atoms, $\forall (k+1 \leq i \leq n)$ $A_i$ are hybrid basic formulae or probability aggregate atoms, and $\forall (1 \leq i \leq n)$ $\mu_i$  are probability annotations.
\\
\\
A DHPP$^{\cal PA}$ rule says that if for each $A_i:\mu_i$, where $k+1 \leq i \leq m$, it is {\em believable} that the probability interval of $A_i$ is at least $\mu_i$ w.r.t. $\leq_t$ and for each $not\; A_j:\mu_j$, where $m+1 \leq j \leq n$, it is \emph{not believable} that the probability interval of $A_j$ is at least $\mu_j$ w.r.t. $\leq_t$, then there exists at least $a_i$, where $1 \leq i \leq k$, such that the probability interval of $a_i$ is at least $\mu_i$.

\begin{definition}
A DHPP$^{\cal PA}$ program over a set of arbitrary p-strategies, $S = S_{conj} \cup S_{disj}$, is a pair $\Pi = \langle R, \tau \rangle$, where $R$ is a set of DHPP$^{\cal PA}$ rules with p-strategies from $S$, and $\tau$ is a mapping $\tau: {\cal B_L} \rightarrow S_{disj}$.
\end{definition}
The mapping $\tau$ in the DHPP$^{\cal PA}$ program definition associates to each atom, $a$, a disjunctive p-strategy that is used to combine the probability intervals obtained from different DHPP$^{\cal PA}$ rules with $a$ appearing in their heads. For the simplicity of the presentation, hybrid basic formulae that appearing in DHPP$^{\cal PA}$ programs without probability annotations are assumed to be associated with the probability annotation $[1,1]$. Nevertheless, probability annotated hybrid basic formulae of the form $A:[P,P]$ are simply represented as $A:P$.

\begin{example} The stochastic dietary problem described in Example (\ref{ex:DHPP_encode}) can be concisely and intuitively represented as DHPP$^{\cal PA}$ program, $\Pi = \langle R, \tau \rangle$, where $\tau$ is any arbitrary assignments of disjunctive p-strategies and $R$ consists of the following DHPP$^{\cal PA}$ rules in addition to the facts represented by $units(F, V, U, S): P$ and $food(X)$ described in Example (\ref{ex:DHPP_encode}).
\[
\begin{array}{lcl}
pckg(F, 1, S) \;\; \vee \;\; pckg(F, 2, S) & \leftarrow & food(F) \\
nutr(F, V, U \times N, S) : P & \leftarrow & units(F, V, U, S) : P, \\ && pckg(F, N, S)
\end{array}
\]
\[
\begin{array}{lcl}
\Gamma  & \leftarrow & not \; \Gamma, \: val_E \{ X : P \: | \: nutr(F, a, X, S):P  \} < 230 \\
\Gamma & \leftarrow & not \; \Gamma, \: val_E \{ X : P \: | \:  nutr(F, b, X, S):P \} < 75 \\
\Gamma  & \leftarrow & not \; \Gamma, \: val_E \{ X : P \: | \: nutr(F, c, X, S):P \} < 95
\end{array}
\]
where the expected value is computed by the probability aggregate $val_E$. The last three DHPP$^{\cal PA}$ rules of the DHPP$^{\cal PA}$ program representation of the stochastic dietary problem described above guarantee that only probability answer sets that involve sufficient daily supply of each vitamin are generated.
\label{ex:main}
\end{example}

\begin{definition}
The {\em ground instantiation} of a symbolic probability set $$S = \{ F:[P_1,P_2]  \; | \; C \}$$ is the set of all ground pairs of the form $\langle \theta\; (F) :[\theta\; (P_1), \theta\; (P_2)]\; | \;  \theta \; (C) \rangle$, where $\theta$ is a substitution of every local variable appearing in $S$ to a constant from $U_{\cal L}$.
\end{definition}

\begin{definition} A ground instantiation of a DHPP$^{\cal PA}$ rule, $r$, is the replacement of each global variable appearing in $r$ to a constant from $U_{\cal L}$, then followed by the ground instantiation of every symbolic probability set, $S$, appearing in $r$.

The ground instantiation of a DHPP$^{\cal PA}$ program, $\Pi$, is the set of all possible ground instantiations of every DHPP$^{\cal PA}$ rule in $\Pi$.
\end{definition}

\begin{example} The ground instantiation of the DHPP$^{\cal PA}$ rule
\[
\Gamma  \leftarrow  not \; \Gamma, \: val_E\{ X : P \: | \: nutr(F, a, X, S):P  \} < 230
\]
with respect to the DHPP$^{\cal PA}$ program, $\Pi$, in Example (\ref{ex:main}), is given by:
{\small
\[
\begin{array}{l}
\Gamma   \leftarrow  not \; \Gamma, \: val_E \{ \\
\langle 60 : 0.7   | nutr(beef, a, 60 , s_1) : 0.7 \rangle ,
\langle 120 : 0.7  | nutr(beef, a, 120, s_1) : 0.7 \rangle, \\
\langle 50 : 0.3   | nutr(beef, a, 50, s_2) : 0.3 \rangle,
\langle 100 : 0.3  | nutr(beef, a, 100, s_2) : 0.3 \rangle,
\\
\langle 8 : 0.8   |  nutr(fish, a, 8 , s_1) : 0.8 \rangle,
\langle 16 : 0.8  |  nutr(fish, a, 16, s_1) : 0.8 \rangle, \\
\langle 11 : 0.2  |  nutr(fish, a, 11, s_2) : 0.2 \rangle,
\langle 22 : 0.2  |  nutr(fish, a, 22, s_2) : 0.2 \rangle,
\\
\langle 60 : 0.8  | nutr(turk, a, 60 , s_1) : 0.8 \rangle,
\langle 120 : 0.8 |  nutr(turk, a, 120, s_1) : 0.8 \rangle, \\
\langle 55 : 0.2  |  nutr(turk, a, 55, s_2) : 0.2 \rangle,
\langle 110 : 0.2 |   nutr(turk, a, 110, s_2) : 0.2 \rangle, \\
\ldots \} < 230
\end{array}
\]
}
\end{example}

\section {Probability Aggregates Semantics}

We present two types of probability aggregates. The first type computes the expected value of a classical aggregate, e.g., the expected value of the minimum, denoted by $f \in \{$ $val_E$, $sum_E$, $times_E$, $min_E$, $max_E$, $count_E$ $\}$, where $val_E$ returns the expected value of a random variable and $sum_E$, $times_E$, $min_E$, $max_E$, $count_E$ return the expected value of the the classical aggregates $sum$, $times$, $min$, $max$, $count$ respectively. The second type of probability aggregates computes the probability of a classical aggregate, e.g, the probability of sum of values, denoted by $g \in \{$ $sum_P$, $times_P$, $min_P$, $max_P$, $count_P$  $\}$, where $sum_P$, $times_P$, $min_P$, $max_P$, $count_P$ return the probability of the the classical aggregates $sum$, $times$, $min$, $max$, $count$ respectively. Any probability aggregate is applied to a probability set that represents a random variable with all its possible values and their associated provability intervals.

\subsection{Probability Aggregates Mappings}

Let $\mathbb{X}$ be a set of objects. Then, we use $2^\mathbb{X}$ to denote the set of all {\em multisets} over elements in $\mathbb{X}$. Let $\mathbb{R}$ denotes the set of all real numbers and $\mathbb{N}$ denotes the set of all natural numbers, and $U_{\cal L}$ denotes the Herbrand universe. Let $\bot$ be a symbol that does not occur in ${\cal L}$. Therefore,

\begin{itemize}

\item The mappings for the expected value probability aggregates are:

\begin{itemize}

\item $val_E :  2^{\mathbb{R} \times {\cal C}[0, 1]} \rightarrow [\mathbb{R}, \mathbb{R}]$.

\item $sum_E :  2^{\mathbb{R} \times {\cal C}[0, 1] } \rightarrow [\mathbb{R}, \mathbb{R}]$.

\item $times_E: 2^{\mathbb{R} \times {\cal C}[0, 1] } \rightarrow [\mathbb{R}, \mathbb{R}]$.

\item $min_E, max_E: (2^{\mathbb{R} \times {\cal C}[0, 1]} - \emptyset) \rightarrow [\mathbb{R}, \mathbb{R}]$.

\item $count_E : 2^{U_{\cal L} \times {\cal C}[0, 1]} \rightarrow [\mathbb{R}, \mathbb{R}]$.

\end{itemize}

\item The mappings for the probability value probability aggregates are:

\begin{itemize}

\item $sum_P : 2^{\mathbb{R} \times {\cal C}[0, 1] } \rightarrow \mathbb{R} \times {\cal C}[0, 1]$.

\item $times_P: 2^{\mathbb{R} \times {\cal C}[0, 1] } \rightarrow \mathbb{R} \times {\cal C}[0, 1]$.

\item $min_P, max_P: (2^{\mathbb{R} \times {\cal C}[0, 1] } - \emptyset) \rightarrow
    \mathbb{R} \times {\cal C}[0, 1]$.

\item $count_P : 2^{U_{\cal L} \times {\cal C}[0, 1]}  \rightarrow \mathbb{N} \times {\cal C}[0, 1]$.

\end{itemize}

\end{itemize}
The application of $sum_E$ and $times_E$ on the empty multiset return $[0,0]$ and $[1,1]$ respectively. The application of $val_E$ and $count_E$ on the empty multiset returns $[0,0]$. The application of $sum_P$ and $times_P$ on the empty multiset return $(0,[1,1])$ and $(1,[1,1])$ respectively. The application of $count_P$ on the empty multiset returns $(0,[1,1])$. However, the application of $max_E$, $min_E$, $max_P$, $min_P$ on the empty multiset is undefined.

\begin{definition}
A probability interpretation, p-interpretation, of a DHPP$^{\cal PA}$ program, $\Pi = \langle R, \tau \rangle$, is a mapping $h: bf_{S}({\cal B_L}) \rightarrow {\cal C}[0, 1]$.
\end{definition}

\subsection{Semantics of Probability Aggregates}

The semantics of probability aggregates is defined with respect to a p-interpretation, which is in turn a representation of probability sets. A probability annotated hybrid basic formula, $A:\mu$, is true (satisfied) with respect to a p-interpretation, $h$, if and only if $\mu \leq_t h(A)$. The negation of a probability annotated hybrid basic formula, $not \; A:\mu$, is true (satisfied) with respect to $h$ if and only if $\mu \nleq_t h(A)$. The evaluation of a probability aggregate, and hence the truth valuation of a probability aggregate atom, are established with respect to a given p-interpretation, $h$, as described by the following definitions.

\begin{definition} Let $f(S)$ be a ground probability aggregate and $h$ be a p-interpretation. Then, we define $S_h$ to be the multiset constructed from elements in the ground $S$, where $S_h = \{\!\!\{ F^g : [P^g_1, P^g_2]  \; | \; \langle F^g : [P^g_1, P^g_2] \; | \; C^g \rangle \in S \wedge$ $C^g$ is true w.r.t. $h \}\!\!\}$.
\end{definition}

\begin{definition} Let $f(S)$ be a ground probability aggregate and $h$ be a p-interpretation. Then, the evaluation of $f(S)$ with respect to $h$ is, $f(S_h)$, the result of the application of $f$ to $S_h$, where $f(S_h) = \bot$ if $S_h$ is not in the domain of $f$ and

\begin{itemize}

\item $val_E(S_h) = \sum_{F^g : [P^g_1, P^g_2] \in S_h} \;( F^g \times [P^g_1, P^g_2])$

\item $sum_E(S_h) = (\sum_{F^g : [P^g_1, P^g_2] \in S_h} \; F^g) \; \times \; X$

\item $times_E(S_h) = (\prod_{F^g : [P^g_1, P^g_2] \in S_h} \; F^g ) \; \times  \; X$

\item $min_E (S_h)= (\min_{F^g : [P^g_1, P^g_2] \in S_h} \; F^g) \; \times \; X$

\item $max_E (S_h)= (\max_{F^g : [P^g_1, P^g_2] \in S_h} \; F^g) \; \times \; X$

\item $count_E (S_h)= (count_{F^g : [P^g_1, P^g_2] \in S_h} \; F^g)  \; \times \; X$
\\

\item $sum_P (S_h) = (\sum_{F^g : [P^g_1, P^g_2] \in S_h} \; F^g \;, \; X)$

\item $times_P (S_h) = (\prod_{F^g : [P^g_1, P^g_2] \in S_h} \; F^g \;, \; X)$

\item $min_P (S_h) = (\min_{F^g : [P^g_1, P^g_2] \in S_h} \; F^g \; , \; X)$

\item $max_P (S_h) = (\max_{F^g : [P^g_1, P^g_2] \in S_h} \; F^g \; , \; X)$

\item $count_P (S_h) = (count_{F^g : [P^g_1, P^g_2] \in S_h} \; F^g \; , \; X)$

\end{itemize}
where $X = \prod_{F^g : [P^g_1, P^g_2] \in S_h} \; [P^g_1, P^g_2]$.
\label{def:ExpProb}
\end{definition}

\section{DHPP$^{\cal PA}$ Probability Answer Set Semantics}

In this section we define the satisfaction, probability models, and the probability answer set semantics of probability aggregates disjunctive hybrid probability logic programs, DHPP$^{\cal PA}$.

Let $r$ be a DHPP$^{\cal PA}$ rule and \\ $head(r) = a_1:\mu_1 \; \vee \ldots \vee \; a_k:\mu_k$ and $body(r) = A_{k+1}:\mu_{k+1}, \ldots, A_m:\mu_m, not\;A_{m+1}:\mu_{m+1},\ldots, not\;A_{n}:\mu_{n}$. We consider that probability annotated probability aggregate atoms that involve probability aggregates from $\{$$val_E$, $sum_E$, $times_E$, $min_E$, $max_E$, $count_E$$\}$ are associated to the probability annotation $[1,1]$.

\begin{definition}
Let $\Pi = \langle R, \tau \rangle$ be a ground DHPP$^{\cal PA}$ program, $r$ be a DHPP$^{\cal PA}$ rule in $R$, $h$ be a p-interpretation for $\Pi$, $f \in \{$$val_E$, $sum_E$, $times_E$, $min_E$, $max_E$, $count_E$$\}$, and $g \in \{$$sum_P$, $times_P$, $min_P$, $max_P$, $count_P$$\}$. Then,

\begin{enumerate}

\item $h$ satisfies $a_i:\mu_i$ in $head(r)$ iff  $\mu_i \leq_t h(a_i)$.

\item $h$ satisfies $f(S) \prec T : [1,1]$ in $body (r)$ iff $f(S_h) \neq \bot$ and $f(S_h) \prec T$.

\item $h$ satisfies $not \; f(S) \prec T :[1,1] $ in $body (r)$ iff $f(S_h) =  \bot$ or  $f(S_h) \neq \bot$ and $f(S_h) \nprec T$.

\item $h$ satisfies $g(S) \prec T : \mu$ in $body (r)$ iff $g(S_h) = (x,\nu) \neq \bot$ and $x \prec T$ and $\mu \leq_t \nu$.

\item $h$ satisfies $not \; g(S) \prec T :\mu $ in $body (r)$ iff $g(S_h) =  \bot$ or $g(S_h) = (x, \nu ) \neq \bot$ and $x \nprec T$ or $\mu \nleq_t \nu$.

\item $h$ satisfies $A_i : \mu_i$ in $body(r)$ iff  $\mu_i \leq_t h(A_i)$.

\item $h$ satisfies $not\;A_j:\mu_j$ in $body(r)$ iff $\mu_j \nleq_t h(A_j)$.

\item $h$ satisfies $body(r)$ iff $\forall(k+1 \leq i \leq m), h$ satisfies $A_i : \mu_i$ and $\forall(m+1 \leq j
\leq n), h$ satisfies $not\;A_j : \mu_j$.

\item $h$ satisfies $head(r)$ iff $\exists i$ $(1 \leq i \leq k)$ such that $h$ satisfies $a_i : \mu_i$.

\item $h$ satisfies $r$ iff $h$ satisfies $head(r)$ whenever $h$ satisfies $body(r)$ or $h$ does not satisfy $body(r)$.

\item $h$ satisfies $\Pi$ iff $h$ satisfies every DHPP$^{\cal PA}$ rule in $R$ and

\begin{itemize}
\item $c_{\tau(a_i)} \{\!\!\{\mu_i \;  \; | \; head(r) \leftarrow body(r) \in R\}\!\!\}\leq_t h(a_i)$ such that $h$ satisfies $body(r)$ and $h$ satisfies \\ $a_i:\mu_i$ in the $head(r)$.

\item $c_\rho \{\!\!\{ h(a_1), \ldots, h(a_n) \}\!\!\} \leq_t h(A)$ such that $a_1, \ldots, a_n$ are atoms in $\cal B_L$ and \\ $A = a_1 *_\rho \ldots *_\rho a_n$ is hybrid basic formula in $bf_S({\cal B_L})$ and $* \in \{\wedge, \vee\}$.
\end{itemize}

\end{enumerate}
\end{definition}

\begin{example} Let $\Pi = \langle R, \tau \rangle$ be a DHPP$^{\cal PA}$ program, where $\tau$ is any arbitrary assignments of disjunctive p-strategies and $R$ consists of the DHPP$^{\cal PA}$ rules:
\[
\begin{array}{lcl}
a(1,1):0.5 \;\; \vee \;\; a(2,1):0.5 & \leftarrow &  \\
a(1, 2):0.7 \;\; \vee \;\; a(2,2):0.3 & \leftarrow &  \\
\end{array}
\]
\[
\begin{array}{lcl}
r: \quad \Gamma  & \leftarrow & not \; \Gamma, \\ && sum_P\{ X : P \: | \: a(X, Y): P  \} \geq 3 \;  : \; 0.3
\end{array}
\]
The ground instantiation of $r$ is given by:
\[
\begin{array}{l}
r': \quad \Gamma   \leftarrow  not \; \Gamma,  \: sum_P\{ \\
\langle 1 : 0.5 \:|\: a(1, 1): 0.5 \rangle, \; \langle 2 : 0.5 \:|\: a(2, 1): 0.5 \rangle, \\
\langle 1 : 0.7 \:|\: a(1, 2): 0.7 \rangle, \; \langle 2 : 0.3 \:|\: a(2, 2): 0.3 \rangle \} \geq 3 \;  : \; 0.3
\end{array}
\]
Let $h$ be a p-interpretation for $\Pi$ that assigns $0.7$ to $a(1,2)$, $0.5$ to $a(2,1)$, and $0$ to the remaining hybrid basic formulae in $bf_{S}({\cal B_L} )$. Thus the evaluation of the probability aggregate atom, $sum_P(S) \geq 3$ in $r'$ w.r.t. to $h$ is given as follows, where
\[
\begin{array}{lcl}
S = \{ &&  \langle 1 : 0.5 \:|\: a(1, 1): 0.5 \rangle, \; \langle 2 : 0.5 \:|\: a(2, 1): 0.5 \rangle, \\&&
\langle 1 : 0.7 \:|\: a(1, 2): 0.7 \rangle, \; \langle 2 : 0.3 \:|\: a(2, 2): 0.3 \rangle \: \}
\end{array}
\]
and $S_h = \{ 1:0.7, \; 2:0.5 \}$. Therefore, $sum_P( \{ 1:0.7, 2:0.5 \} ) = (3, 0.35)$, and consequently, the probability annotated probability aggregate atom $sum_P(S) \geq 3 \;  : \; 0.3$ is satisfied by $h$. This is because $sum_P( \{ 1:0.7, 2:0.5 \} ) = (3, 0.35) \neq \bot$ and $3 \geq 3$ and $0.3 \leq_t 0.35$
\label{ex:dice}
\end{example}
Let $L$ denotes a probability annotated hybrid basic formula, $A:\mu$ or the negation of $A:\mu$, denoted by $not \; A:\mu$. Let $h_1, h_2$ be two p-interpretations. Then, we say that $L$ is monotone if $\forall (h_1, h_2)$ such that $h_1 \leq_t h_2$, it is the case that if $h_1$ satisfies $L$ then $h_2$ also satisfies $L$. However, $L$ is antimonotone if $\forall (h_1, h_2)$ such that $h_1 \leq_t h_2$ it is the case that if $h_2$ satisfies $L$ then $h_1$ also satisfies $L$. But, if $L$ is not monotone or not antimonotone, then we say $L$ is nonmonotone. A probability annotated atom or a probability annotated probability aggregate atom, $a:\mu$, or the negation of probability annotated atom or the negation of a probability annotated probability aggregate atom, $not \; a:\mu$, can be monotone, antimonotone or nonmonotone, since their probability annotations are allowed to be arbitrary functions. Moreover, probability aggregate atoms by themselves can be monotone, antimonotone or nonmonotone. This also carry over to probability annotated hybrid basic formulae.

\begin{definition} A probability model, \emph{p-model}, for a DHPP$^{\cal PA}$ program, $\Pi$, is a p-interpretation for $\Pi$ that satisfies $\Pi$. A p-model $h$ for $\Pi$ is $\leq_t$--minimal iff there does not exist a p-model $h^\prime$ for $\Pi$ such that $h^\prime <_t h$.
\end{definition}

\begin{example} It can easily verified that the p-interpretation, $h$, for DHPP$^{\cal PA}$ program, $\Pi$, described in Example (\ref{ex:dice}), is not a p-model for $\Pi$. However, by considering only the relevant hybrid basic formulae, the following p-interpretation, $h'$, is a p-model for $\Pi$, where $ h'=   \{ a(1,1):0.5, \; a(1, 2):0.7, \ldots \}$.
\end{example}

\begin{definition} Let $\Pi = \langle R, \tau \rangle$ be a ground DHPP$^{\cal PA}$ program, $r$ be a DHPP$^{\cal PA}$ rule in $R$, and $h$ be a p-interpretation for $\Pi$. Let $h \models body(r)$ denotes $h$ satisfies $body(r)$. Then, the probability reduct, $\Pi^h$, of $\Pi$ w.r.t. $h$ is a ground DHPP$^{\cal PA}$ program $\Pi^h = \langle R^h, \tau \rangle$ where
\[
R^h = \{ head(r) \leftarrow body(r) \: \: | \: \: r \in R \: \wedge \: h \models body(r)\}
\]
\end{definition}

\begin{definition} A p-interpretation, $h$, of a ground DHPP$^{\cal PA}$ program, $\Pi$, is a probability answer set for $\Pi$ if $h$ is $\leq_t$-minimal p-model for $\Pi^h$.
\end{definition}
Observe that the definitions of the probability reduct and the probability answer sets for DHPP$^{\cal PA}$ programs are generalizations of the probability reduct and the probability answer sets of the original DHPP programs described in \cite{Saad_DHPP}.

\begin{example} It can be easily verified that the DHPP$^{\cal PA}$ program presented in Example (\ref{ex:dice}) has three probability answer sets, which by considering relevant hybrid basic formulae are:
\[
\begin{array}{lcl}
h_1 & = &  \{ a(1,1):0.5, \; a(1, 2):0.7, \ldots \} \\
h_2 & = &  \{ a(1,1):0.5, \; a(2,2):0.3, \ldots \} \\
h_3 & = &  \{ a(2,1):0.5, \; a(2,2):0.3, \ldots \}
\end{array}
\]
\end{example}

\begin{example} The stochastic dietary problem representation by the DHPP$^{\cal PA}$ program described in Example (\ref{ex:main}) has four probability answer sets, which are:
%
%
\[
\begin{array}{l}
h_1 = \{ \:\:\:
pckg(beef,2,s1),
pckg(fish,2,s1),
\\
pckg(turk,2,s1),
pckg(beef,1,s2),
pckg(fish,2,s2),
\\
pckg(turk,2,s2),
nutr(beef,a,120,s1):0.7,
\\
nutr(fish,a,16,s1):0.8,
nutr(turk,a,120,s1):0.8,
\\
nutr(beef,a,50,s2):0.3,
nutr(fish,a,22,s2):0.2,
\\
nutr(turk,a,110,s2):0.2,
nutr(turk,b,30,s1):0.7,
\\
nutr(fish,b,30,s1):0.5,
nutr(beef,b,20,s1):0.6,
\\
nutr(turk,b,40,s2):0.3,
nutr(fish,b,36,s2):0.5,
\\
nutr(beef,b,8,s2):0.4,
nutr(beef,c,40,s1):0.8,
\\
nutr(fish,c,26,s2):0.6,
nutr(turk,c,50,s2):0.1,
\\
nutr(beef,c,15,s2):0.2,
nutr(turk,c,40,s1):0.9,
\\
nutr(fish,c,20,s1):0.4,
\ldots
\}
\end{array}
\]
\[
\begin{array}{l}
h_2 = \{ \:\:\:
pckg(beef,2,s1),
pckg(fish,2,s1),
\\
pckg(turk,2,s1),
pckg(beef,2,s2),
pckg(fish,2,s2),
\\
pckg(turk,2,s2),
nutr(beef,a,120,s1):0.7,
\\
nutr(fish,a,16,s1):0.8,
nutr(turk,a,120,s1):0.8,
\\
nutr(beef,a,100,s2):0.3,
nutr(fish,a,22,s2):0.2,
\\
nutr(turk,a,110,s2):0.2,
nutr(beef,b,20,s1):0.6,
\\
nutr(fish,b,30,s1):0.5,
nutr(turk,b,30,s1):0.7,
\\
nutr(beef,b,16,s2):0.4,
nutr(fish,b,36,s2):0.5,
\\
nutr(turk,b,40,s2):0.3,
nutr(beef,c,40,s1):0.8,
\\
nutr(fish,c,20,s1):0.4,
nutr(turk,c,40,s1):0.9,
\\
nutr(beef,c,30,s2):0.2,
nutr(fish,c,26,s2):0.6,
\\
nutr(turk,c,50,s2):0.1,
\ldots
\}
\end{array}
\]
\[
\begin{array}{l}
h_3 = \{ \:\:\:
pckg(beef,2,s1),
pckg(fish,2,s1),
\\
pckg(turk,2,s1),
pckg(beef,2,s2),
pckg(fish,2,s2),
\\
pckg(turk,1,s2),
nutr(beef,a,120,s1):0.7,
\\
nutr(fish,a,16,s1):0.8,
nutr(turk,a,120,s1):0.8,
\\
nutr(beef,a,100,s2):0.3,
nutr(fish,a,22,s2):0.2,
\\
nutr(turk,a,55,s2):0.2,
nutr(beef,b,20,s1):0.6,
\\
nutr(fish,b,30,s1):0.5,
nutr(turk,b,30,s1):0.7,
\\
nutr(beef,b,16,s2):0.4,
nutr(fish,b,36,s2):0.5,
\\
nutr(turk,b,20,s2):0.3,
nutr(beef,c,40,s1):0.8,
\\
nutr(fish,c,20,s1):0.4,
nutr(turk,c,40,s1):0.9,
\\
nutr(beef,c,30,s2):0.2,
nutr(fish,c,26,s2):0.6,
\\
nutr(turk,c,25,s2):0.1,
\ldots
\}
\\
\\
h_4 =  \{ \:\:\:
pckg(beef,2,s1),
pckg(fish,1,s1),
\\
pckg(turk,2,s1),
pckg(beef,2,s2),
pckg(fish,2,s2),
\\
pckg(turk,2,s2),
nutr(beef,a,120,s1):0.7,
\\
nutr(fish,a,8,s1):0.8,
nutr(turk,a,120,s1):0.8,
\\
nutr(beef,a,100,s2):0.3,
nutr(fish,a,22,s2):0.2,
\\
nutr(turk,a,110,s2):0.2,
nutr(beef,b,20,s1):0.6,
\\
nutr(fish,b,15,s1):0.5,
nutr(turk,b,30,s1):0.7,
\\
nutr(beef,b,16,s2):0.4,
nutr(fish,b,36,s2):0.5,
\\
nutr(turk,b,40,s2):0.3,
nutr(beef,c,40,s1):0.8,
\\
nutr(fish,c,10,s1):0.4,
nutr(turk,c,40,s1):0.9,
\\
nutr(beef,c,30,s2):0.2,
nutr(fish,c,26,s2):0.6,
\\
nutr(turk,c,50,s2):0.1,50,
\ldots
\}
\end{array}
\]
%
\end{example}

\section{DHPP$^{\cal PA}$ Semantics Properties}

In this section we study the semantics properties of DHPP$^{\cal PA}$ programs and its relationship to the original probability answer set semantics of disjunctive hybrid probability logic programs, denoted by DHPP \cite{Saad_DHPP}; the classical answer set semantics of classical disjunctive logic programs with classical aggregates, denoted by DLP$^{\cal A}$ \cite{Recur-aggr}; and the original classical answer set semantics of classical disjunctive logic programs, denoted by DLP \cite{Gelfond_B}.

\begin{theorem} Let $\Pi$ be a DHPP$^{\cal PA}$ program. The probability answer sets for $\Pi$ are $\leq_t$--minimal p-models for $\Pi$.
\end{theorem}
The following theorem shows that the probability answer set semantics of DHPP$^{\cal PA}$ programs subsumes and generalizes the probability answer set semantics of DHPP \cite{Saad_DHPP} programs, which are DHPP$^{\cal PA}$ programs without probability aggregate atoms and with only monotone probability annotation functions.

\begin{theorem} Let $\Pi$ be a DHPP program and $h$ be a p-interpretation. Then, $h$ is a probability answer set for $\Pi$ iff $h$ is a probability answer set for $\Pi$ w.r.t. the probability answer set semantics of \cite{Saad_DHPP}.
\end{theorem}
In what follows we show that the probability answer set semantics of DHPP$^{\cal PA}$ programs naturally subsumes and generalizes the classical answer set semantics of the classical disjunctive logic programs with the classical aggregates, DLP$^ {\cal A}$ \cite{Recur-aggr}, which consequently naturally subsumes the classical answer set semantics of the original classical disjunctive logic programs, DLP \cite{Gelfond_B}.

Any DLP$^{\cal A}$ program, $\Pi$, is represented as a DHPP$^{\cal PA}$ program, $\Pi^\prime = \langle R, \tau \rangle$, where each DLP$^{\cal A}$ rule in $\Pi$ of the form
\[
a_1 \; \vee \ldots \vee \; a_k \leftarrow a_{k+1}, \ldots, a_m, not\; a_{m+1},\ldots, not \;a_{n}
\]
is represented, in $R$, as a DHPP$^{\cal PA}$ rule of the form
\[
\begin{array}{r}
a_1:[1,1] \; \vee \ldots \vee \; a_k:[1,1] \leftarrow a_{k+1}:[1,1], \ldots, a_m:[1,1], \\ not\; a_{m+1}:[1,1],\ldots, not \;a_{n}:[1,1]
\end{array}
\]
where $a_1, \ldots, a_k$ are atoms and $a_{k+1},\ldots, a_n$ are atoms or probability aggregate atoms whose probability aggregates contain probability sets that involve conjunctions of probability annotated atoms with probability annotation $[1,1]$, where $[1,1]$ represents the truth value \emph{true}. In addition, $\tau$ is any arbitrary assignments of disjunctive p-strategies. We call this class of DHPP$^{\cal PA}$ programs as DHPP$_1 ^{\cal PA}$ programs. Any DLP program is represented as a DHPP$_1 ^{\cal PA}$ program by the same way as DLP$^{\cal A}$ except that DLP disallows classical aggregate atoms. The following results show that DHPP$_1 ^{\cal PA}$ programs subsume both DLP$^{\cal A}$ and DLP programs.

\begin{theorem} Let $\Pi'$ be the DHPP$_1 ^{\cal PA}$ program equivalent to a DLP$^{\cal A}$ program $\Pi$. Then, $h$ is a probability answer set for $\Pi'$ iff $I$ is a classical answer set for $\Pi$, where $h(a) = [1,1]$ iff $a \in I$ and $h(b) = [0,0]$ iff $b \in {\cal B_L} - I$.
\label{thm:DLP2DHPP}
\end{theorem}

\begin{proposition} Let $\Pi'$ be the DHPP$_1 ^{\cal PA}$ program equivalent to a DLP program $\Pi$. Then, $h$ is a probability answer set for $\Pi'$ iff $I$ is a classical answer set for $\Pi$, where $h(a) = [1,1]$ iff $a \in I$ and $h(b) = [0,0]$ iff $b \in {\cal B_L} - I$.
\end{proposition}

\section{Conclusions and Related Work}

We presented DHPP$^{\cal PA}$ that extends the original language of DHPP with arbitrary probability annotations functions and arbitrary probability aggregate functions that determine the expected value of the classical aggregate functions and the probability of a classical aggregate functions. We introduced the probability answer set semantics of DHPP$^{\cal PA}$ with arbitrary probability aggregates including monotone, antimonotone, and nonmonotone probability aggregates. We have shown that the DHPP$^{\cal PA}$ probability answer set semantics generalize DHPP original probability answer set semantics \cite{Saad_DHPP}. In addition, we proved that the probability answer sets of DHPP$^{\cal PA}$ are minimal probability models and consequently incomparable, which is an important property for nonmonotonic probability reasoning.

To the best of our knowledge, this development is the first in probability logic programming literature to consider probability aggregates in probability logic programming in general and probability answer set programming in particular. Nevertheless, classical aggregates were extensively investigated in classical answer set programming \cite{Recur-aggr,Smodels-Weight,WFM-Pref,Dis-mono-aggr,Ferraris,FOL-aggr,Pelov}.

Among these investigations, \cite{Recur-aggr} is the most general intuitive semantics for classical aggregates in DLP, since it is  declarative classical answer semantics for classical disjunctive logic program with arbitrary classical aggregates (DLP$^{\cal A}$), including monotone, antimonotone, and nonmonotone aggregates, and a natural generalization of the classical answer set semantics of aggregate-free DLP \cite{Gelfond_B}. We have shown that the probability answer set semantics of DHPP$^{\cal PA}$ subsumes both DLP$^{\cal A}$ and DLP classical answer set semantics. Extensive comparisons between DLP$^{\cal A}$ and the existing approaches to classical aggregates can be fount in \cite{Recur-aggr}. Among these approaches, \cite{Smodels-Weight} that allows only classical aggregates of the form $sum$ and $count$, however, they do not behave intuitively with negative values \cite{Ferraris}. In addition, \cite{FOL-aggr} presented classical aggregates for first-order formulae.

On the other hand probability aggregates are studied in probability databases in the context of query evaluation over probability data \cite{ProbAgg,ProbOLAP,havingPDB}. Two main approaches are available for defining the semantics of probability aggregate queries in probability databases. The first approach adopted in \cite{ProbAgg,ProbOLAP,havingPDB}, applied to OLAP applications, defines the semantics of the probability aggregates queries as the expected value of the aggregate queries over the possible worlds of the probability database. However, the second approach \cite{havingPDB}, defines the semantics of the probability aggregates queries as the probability of the aggregate queries over the possible worlds of the probability database. The possible world semantics is adopted in defining the semantics of probability aggregate queries in both approaches in \cite{ProbAgg,ProbOLAP,havingPDB}. In DHPP$^{\cal PA}$, we considered the two approaches, where probability aggregates are evaluated with respect to a probability answer set, which is considered evaluation over a possible world.

\bibliographystyle{named}
\bibliography{Saad13PAPASP}

\end{document}